\documentclass[10pt,twocolumn,letterpaper]{article}

\usepackage{iccv}
\usepackage{times}
\usepackage{epsfig}
\usepackage{graphicx}
\usepackage{amsmath}
\usepackage{amssymb}
\usepackage{times}
\usepackage{epsfig}
\usepackage{graphicx}
\usepackage{amsmath}
\usepackage{amssymb}
\usepackage{mathrsfs}
\usepackage{multirow}
\usepackage{booktabs}
\newtheorem{lemma}{Lemma}
\usepackage{algorithm}
\usepackage{algorithmic}
\usepackage{bm}
\usepackage{color}

\usepackage[pagebackref=true,breaklinks=true,letterpaper=true,colorlinks,bookmarks=false]{hyperref}

\iccvfinalcopy % *** Uncomment this line for the final submission

\def\iccvPaperID{4} % *** Enter the ICCV Paper ID here

\ificcvfinal\fi

\begin{document}

%%%%%%%%% TITLE
\title{Instance-weighted Central Similarity for Multi-label Image Retrieval}

\author{Zhiwei Zhang\footnotemark[1]\\
Centre for Perceptual and Interactive Intelligence\\
%{\tt\small zwzhang@cpii.hk}
\and
Hanyu Peng\footnotemark[1]\\
Chinese Academy of Sciences\\
}

\maketitle
\ificcvfinal\thispagestyle{empty}\fi

%%%%%%%%% ABSTRACT
\begin{abstract}
   Deep hashing has been widely applied to large-scale image retrieval by encoding high-dimensional data points into binary codes for efficient retrieval. Compared with pairwise/triplet similarity based hash learning, central similarity based hashing can more efficiently capture the global data distribution. For multi-label image retrieval, however, previous methods only use multiple hash centers with equal weights to generate one centroid as the learning target, which ignores the relationship between the weights of hash centers and the proportion of instance regions in the image. To address the above issue, we propose a two-step alternative optimization approach, Instance-weighted Central Similarity (ICS), to automatically learn the center weight corresponding to a hash code. Firstly, we apply the maximum entropy regularizer to prevent one hash center from dominating the loss function, and compute the center weights via projection gradient descent. Secondly, we update neural network parameters by standard back-propagation with fixed center weights. More importantly, the learned center weights can well reflect the proportion of foreground instances in the image. Our method achieves the state-of-the-art performance on the image retrieval benchmarks, and especially improves the mAP by 1.6\%-6.4\% on the MS COCO dataset.
\end{abstract}

\renewcommand{\thefootnote}{\fnsymbol{footnote}}
	\footnotetext[1]{Equal Contribution.}
	\footnotetext[2]{Corresponding author.}

\section{Introduction}

\begin{figure}[t]
\centering
\includegraphics[width=\linewidth]{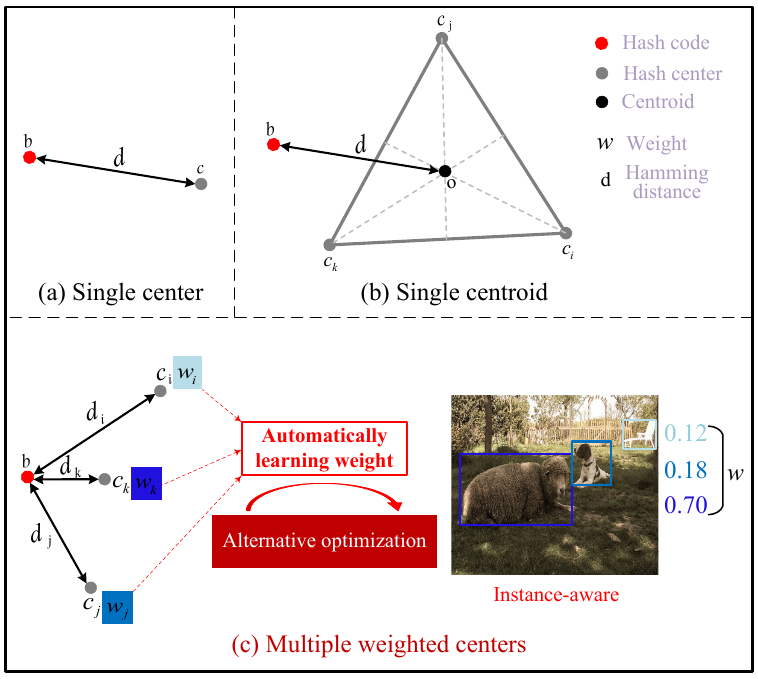}
\caption{Central Similarity based hashing methods for image retrieval. Hash centers are sampled from the row or column vectors of Hadamard matrix. (a) A single center is used as the target to learn hash function for single-label image retrieval. (b) Use multiple hash centers with the equal weights to generate a centroid as the learning target, which does not consider the relationship between the center weights and the proportion of foreground instances in the image. (c) We propose instance-aware weighting approach to automatically learn the hash center weights by an alternative optimization. As illustrated in sub-figure (c), the center weights corresponding to the three foreground instances (shape, dog and chair) are 0.70, 0.18, and 0.12 respectively, which well reflect their proportion in the image.}
\label{fig:motivation}
\end{figure}

Representing images efficiently is an important task for large-scale multimedia retrieval. Hashing has attracted extensive attention due to its computational and storage efficiency. It aims to encode high-dimensional data points into compact binary codes, hence the original feature space can be efficiently approximated via low-dimension Hamming space~\cite{survey2017,survey2020}. Recently, deep learning to hash methods can perform simultaneous feature learning and hash code learning in an end-to-end way, which has shown state-of-the-art performance on many benchmarks~\cite{CNNH2014,DNNH2015,DHN2016,hashnet2017,DCH2018,CSQ2020}.

The existing data-dependent hashing methods mostly generate similar binary codes for similar data points by utilizing pairwise or triplet data similarity~\cite{pairwise2016,wang2016deep,hashnet2017}. Yuan et al.~\cite{CSQ2020} analyzed the limitations of the pairwise/triplet based hashing methods, and proposed the Central Similarity Quantization (CSQ) for image retrieval that not only captures the data similarity globally but also improves the learning efficiency. The existing central similarity methods utilize the row vectors or column vectors of Hadamard matrix to generate hash centers~\cite{HCOH2018,HCDH2019,HMOH2020,CSQ2020}. As shown in Figure \ref{fig:motivation} (a), a single center is used as the target to learn the hash function for single-label image retrieval. For multi-label image retrieval illustrated in \ref{fig:motivation} (b), however, Yuan et al.~\cite{CSQ2020} only uses multiple hash centers with equal importance to generate one centroid as the learning target. The central similarity based methods neglect the relationship between the importance of the hash centers and the proportion of foreground objects in the image. In addition, the above methods only consider the semantic similarities at image level, which may be suboptimal for multi-label image retrieval.

 %Lin et al.~\cite{HMOH2020} also merges the multiple hash centers into one binary code based on ``majority principle'' and ``balanced principle''.

The existing multi-label image retrieval methods can be divided into two main
categories: one is to use the region proposal module to extract foreground objects and generate  multiple hash codes for a multi-lable image~\cite{IAH2016,OLAH2018,Region-DH2020}. The other is to consider the number of common labels of multi-label image as the metric of similarity~\cite{DSRH2015,RCDH2018}. The former methods achieve instance-aware retrie, but their limitation is that the generation of hash codes is dependent on the quality of region proposal module. Therefore, how to make the model automatically learn the multi-scale semantic information contained in the image without using region proposal module?

%There is also some methods using Graph Convolution Network (GCN) for multi-label image retrieval

In this paper, we propose an nstance-weighted Central Similarity (ICS) to learn the relative importance of each hash center corresponding to a hash code as shown in \ref{fig:motivation} (c). Our method use multiple hash centers to guide the model to focus only on the foreground objects and ignore the background. ICS is a two-step alternative optimization approach. In the first step, we apply the maximum entropy regularizer to prevent one hash center from dominating the loss function, and compute the center weights via projection gradient descent. In the second step, the neural network parameters are updated by standard back-propagation. In addition, the center weights learned by ICS can reflect the proportion of foreground objects in the image, instead of leveraging the region proposal generation model to produce hash codes for foreground objects. The performance of proposed ICS for multi-label image retrieval is evaluated on NUS-WIDE~\cite{nus-wide2009} and MS COCO~\cite{coco2014} datasets. Our contributions are summarized as follows.

\begin{itemize}
	\item We first consider the relationship between the importance of hash centers and the proportion of foreground instance regions in the image, and propose the ICS to automatically learn the hash center weights by an alternative optimization.
		
	\item The maximum entropy regularizer in ICS leads the distribution of center weights to be more reasonable and the learning process of center weights can converge within few iterations via project gradient descent.
	
	\item The proposed ICS is also a instance-aware method that the center weight value can well reflect the proportion of foreground objects in the image. Our visualization results can support this point well.
		
	\item Our experimental results demonstrate that the ICS can improve the performance on the three data sets NUS-WIDE, MS COCO and FLICKR25K by 0.4\%-1.1\%, 1.6\%-6.4\% and 6.0\%-7.6\% in mAP, respectively.
\end{itemize}

\section{Related Work}
In this section, we briefly review the related works in terms of deep learning to hash methods, and divide them into two categories according to whether the multi-label information in the image is considered. 

The first category is represented by HashNet~\cite{hashnet2017} and CSQ~\cite{CSQ2020}. 1) HashNet~\cite{hashnet2017} considered the imbalanced similarity relationships between positive and negative pairwise samples, and used weighted maximum likelihood objective function to learn hash codes. DCH~\cite{DCH2018} used Cauchy distribution to improve search performance from a Bayesian perspective. However, the above pairwise/triplet based methods ignore the global structure of samples and suffer from low retrieval efficiency. 2) Recent deep hashing methods~\cite{HCOH2018,lin2019towards,HCDH2019,CSQ2020} based on Hadamard matrix show promising results due to its orthogonal and balanced structure. CSQ~\cite{CSQ2020} leveraged Hadamard matrix to generate hash centers for learning central similarities between samples. HCDH~\cite{HCDH2019} used label information to guide the learning of hash centers. For multi-label images, however, the previous methods ignore the relative importance of each hash center and consider each lable equally in a image. In this paper, we consider the proportion of each foreground object should be different and propose to learn the center weights automatically.

The second category is represented by IAH~\cite{IAH2016} and RCDH~\cite{RCDH2018}. DSRH~\cite{DSRH2015} and IAH~\cite{IAH2016} used the region proposal module to generate multiple hash codes for a multi-lable image. The limitation is that the generation of hash codes is dependent on the quality of region proposal module. RCDH~\cite{RCDH2018} considers the number of common labels of multi-label image as the metric of similarity.

\section{The Proposed Method}
In this section, we first introduce how to generate hash centers via Hadamard matrix. Afterwards,  maximum entropy regularizer is applied to learn the relative importance of each hash center. Finally, we propose the two-step based optimization approach to update the hash center weights and neural network parameters alternatively.

\subsection{Problem Definition}
Let $\mathcal{X}$ be a set of training data, $i.e$. $\mathcal{X} = \left\{\left\{\mathbf{x}_i\right\}_{i=1}^N \in \mathbb{R}^D, \left\{\mathbf{y}_i\right\}_{i=1}^{N} \in \mathbb{R}^M \right\}$, where $D$ is the dimension of input, $M$ denotes the number of labels, and $\mathbf{y}_i$ stands for the label corresponding to the $i$-th sample. Our goal is to learn a nonlinear hash function $f_{\theta}: \mathbf{x} \mapsto \mathbf{b} \in \left\{-1, 1\right\}^K$ that encodes each data point $\mathbf{x}$ into compact $K$-bit hash code, where $\theta$ denotes the model parameters. At the same time, similarities in the original space are preserved in the Hamming space. Given a set of \textit{hash centers} $\mathcal{V} = \left\{\mathbf{v}_1, \mathbf{v}_2, ..., \mathbf{v}_m\right\} \in \left\{-1, 1\right\}^K$ for multi-label image retrieval, each binary code $\mathbf{b}_i$ has hash centers $\left\{\mathbf{v}_{ij}\right\}_{j=1}^{c_i}$ and their corresponding weights $\mathbf{w}_i=\left\{{w}_{ij}\right\}_{j=1}^{c_i}$, where $c_i$ represents the number of labels of $\mathbf{x}_i$. We aim to automatically learn the hash center weights $\mathbf{w}$. In addition, we hope that the value of $\mathbf{w}$ can reflect the proportion of the corresponding foreground objects in the image.

\subsection{Generation of Hash Centers}
Following ~\cite{CSQ2020}, we define the row vectors in Hadamard matrix as hash centers. Hadamard matrix satisfies the independence and balance principles that its row vectors and column vectors are pair-wise orthogonal, and half of bits are 1 or -1. The generation of $2^{k}$-order Hadamard matrix $H$ is rather simple using the Sylvester’s algorithm~\cite{hadamard1867}:
\begin{equation}
\begin{split}
H_{2^{k}} &= \left[ \begin{array}{cc}
H_{2^{k-1}} & H_{2^{k-1}} \\
H_{2^{k-1}} & -H_{2^{k-1}} 
\end{array} \right] = H_{2} \otimes H_{2^{k-1}}, \\
H_{2}& = \left[ \begin{array}{cc}
1 & 1 \\
1 & -1 
\end{array} \right],
\end{split}
\label{eq-hadamard}
\end{equation}
where $\otimes$ represents the Kronecker product.

The generation of hash centers is related to the number hash bits $K$ and the number of class labels $M$, which includes three strategies: 1) When $M \leq K$, the row vectors are randomly sampled from $H$ as hash center. 2) When $K < M \leq 2K$, a new Hadamard matrix $H_{2 \cdot 2^{k}}$ = $[H_{2^{k}}, -H_{2^{k}}]^{\top}$ is utilized to sample hash centers. 3) When $2K \geq M$ and $K$ is not the power of 2, each bit in a hash center $v$ is sampled from a Bernoulli distribution Bern(0.5), which means that random half of bits are 1 or -1.

\subsection{Instance-weighted Central Similarity}
In order to measure the similarity of hash code $\{\mathbf{b}_i\}_{i=1}^N$ and its corresponding hash centers $\{\mathbf{v}_{ij}\}_{j=1}^{c_i}$, Binary Cross Entropy (BCE) is used to measure the Hamming distance $d_{i}$ between the hash code $\mathbf{b}_{i}$ and its hash centers $\mathbf{v}_{i}$. The distance function $d_{i} = \Omega (\mathbf{b}_i, \mathbf{v}_{i})$ can be written as: 
\begin{equation}
\begin{split}
d_{i} =& \Omega (\mathbf{b}_i, \mathbf{v}_{i}) \\
=& \sum_{j=1}^{c_i} \sum_{k=1}^K (v_{ij}^{k} log b_{i}^{k} + (1-v_{ij}^{k})log(1-b_{i}^{k})),
\end{split}
\label{eq2}
\end{equation}
As illustrated in Figure \ref{fig:motivation}, the proportion of foreground instances in the image is different. Therefore, for the hash centers $\{\mathbf{v}_{ij}\}_{j=1}^{c_i}$ representing different labels, hash code $\mathbf{b}_i$ should be given different weights according to the distance.
%the learned hash code relative importance of each hash center should also be taken into consideration. Inspired by this, 
We think the Hamming distance between each hash code $\mathbf{b}_i$ and its corresponding hash centers $\{v_{ij}\}_{j=1}^{c_i}$ should be weighted non-uniformly. We introduce weights $\{w_{ij}\}_{j=1}^{c_i}$ to reflect the weights of hash centers and assume that it is learnable rather than with the same importance. Therefore, we design a weighted central similarity loss:
\begin{equation}
\label{eq3}
\begin{split}
    \Omega(\mathbf{b}_i, \mathbf{v}_{i}) = \sum_{j=1}^{c_i} w_{ij} d_{ij},\\
    s.t. \sum_{j=1}^{c_i} w_{ij}=1, 
\end{split}
\end{equation}
where $j = 1,2, ..., c_i$ represents the number of hash centers corresponding to a hash code. 
%In this sense, the above formula becomes a general framework for Hamming distance. 
In addition, we regard the probabilities of assigning each hash code to the hash centers as a classification problem. We convert the distance $\Omega(\mathbf{b}_i, \mathbf{v}_{i})$ based on the \textit{adaptive} Sigmoid function as follows: 
\begin{equation}
\label{eq4}
p(\Omega(\mathbf{b}_i,\mathbf{v}_{i}))=\sigma(-\Omega(\mathbf{b}_i,\mathbf{v}_{i}))=\frac{1}{1+e^{\beta \Omega(\mathbf{b}_i,\mathbf{v}_{i})}},
\end{equation}
where $\beta$ is a smooth parameter to control its bandwidth. Inspired from HashNet~\cite{hashnet2017}, the Sigmoid function with larger $\beta$ will have larger saturation zone where its gradient is zero. To perform more effective back-propagation, we usually require $\beta \leq$ 1 in our experiments. It is easy to see that the smaller $\Omega(\mathbf{b}_i, \mathbf{v}_{i})$ is,  the larger $p(\Omega(\mathbf{b}_i,\mathbf{v}_i))$ will be. Therefore, maximizing the likelihood of $p(\Omega(\mathbf{b}_i,\mathbf{v}_{i}))$ will make the Hamming distance between hash centers and hash codes small, which means that the foreground instances with larger proportion in an image, the higher classification probability of the corresponding hash centers. 

We define the following likelihood function to learn center weight $\mathbf{w}$:
\begin{equation}
\label{eq5.1}
L(\mathbf{\theta})=\prod_{i=1}^{N}p(\Omega(\mathbf{b}_i,\mathbf{v}_{i});\theta),
\end{equation}\label{eq5.2}
By taking log of $p(\Omega(\mathbf{b}_i,\mathbf{v}_i))$, it can be derived as:
\begin{equation}
-\log(L(\mathbf{\theta}))=-\sum_{i=1}^{N}\log(p(\Omega(\mathbf{b}_i,\mathbf{v}_i);\theta)),
\end{equation}\label{eq5.3}

We redefine the loss function as:
\begin{equation}
\min_{\mathcal{B}} \mathcal{J}_1 = -\sum_{i=1}^{N}\log(p(\Omega(\mathbf{b}_i,\mathbf{v}_{i});\theta)),
\end{equation}\label{eq5.4}
We can expand this formula as:
\begin{equation}
\label{eq6}
\begin{aligned}
\min_{\mathcal{B}} \mathcal{J}_1 =& \sum_{i=1}^N\beta(\sum_{j=1}^{c_i}\log(1+\exp(w_{ij}(\sum_{k=1}^K [v^i_{j,k}\log b_{i,k} \\ 
+& (1-v_{j,k}^i)\log(1-b_{i,k}) ])) + w_{ij}(\sum_{k=1}^K  [v^i_{j,k}\log b_{i,k} \\
+& (1-v_{j,k}^i)\log(1-b_{i,k})  ])).
\end{aligned}
\end{equation}
However, the direct minimization of above equation with respect to $w_{ij}$ may produce unreasonable hash center weights. This is obvious from the following Lemma~\ref{lemma}.

\begin{lemma}
\label{lemma}
Set $d_{i \star}=min_{1\le j \le c_i}d_{ij}$, then $ \Omega(\mathbf{b}_i, \mathbf{v}_{i}) = \sum_{j=1}^{c_i}w_{ij}d_{ij}\ge d_{i \star}$.
\end{lemma}

We can obtain a unique minimum solution for Eq.~\ref{eq6} with the weights $w_{ij}=1$ if $j=argmin_{1\le j \le c_i}d_{ij}$ and 0 elsewhere. However, the weighted distance function for a multi-label image only minimizes Hamming distance between a hash code and its corresponding one hash center, which obviously makes no sense for multi-label image retrieval. To prevent one hash center from dominating the loss function, we apply the maximum entropy principle to learn the hash center weights. 
\begin{equation}
\label{eq8}
R(w_i) = \sum_{j=1}^{c_i}w_{ij}log(w_{ij}),
\end{equation}

We can reformulate the loss function as :
\begin{equation}
\centering
\begin{split}
    \mathcal{J}=\mathcal{J}_1  + \lambda R,\\
    s.t. \sum_{j=1}^{c_i} w_{ij}=1, \\
     w_{ij} \geq 0.
\end{split}
\label{eq8.0}
\end{equation}

The above equation has a new parameter $\lambda$ that is used to control the degree of maximum cross entropy regularization. If $\lambda$ approaches zeros from positive and $w_{ij}$ approaches $\frac{1}{c_i}$, then the minimization of above loss function tends to give equal weight to each hash center. If $\lambda$ approaches a positive infinity, then $w_{ij}$ will approach $w_{ij}=1$ if $j=argmin_{1\le j \le c_i}d_{ij}$ and 0 elsewhere. Obviously, such a result is not suitable for multi-label image retrieval. The detailed analysis of how to choose a proper $\lambda$ will be further demonstrated in section~\ref{sec:experiment}.

We also use the quantization loss $\mathcal{J}_q$ to make the hash codes converge to hash centers~\cite{hyvarinen2009natural}. 
\begin{equation}
\label{eq7}
\mathcal{J}_q = \sum_{i=1}^N \sum_{k=1}^K (log \, cosh(|2b_{i,k}-1 | - 1)).
\end{equation}
Our final optimization function is:
\begin{equation}
\centering
\label{eq9}
\begin{split}
    \mathcal{J}=\mathcal{J}_1 +\gamma \mathcal{J}_q + \lambda R,\\
    s.t. \sum_{j=1}^{c_i} w_{ij}=1, \\
     w_{ij} \geq 0.
\end{split}
\end{equation}
where $\gamma$ is fixed and set to 0.05.

\begin{algorithm}[tb]
\caption{Instance-weighted Central Similarity}
		
\textbf{Input:} Training data $\mathcal{X} = \left\{\mathbf{x}_i\right\}_{i=1}^N \in \mathbb{R}^D$, hash centers $\mathcal{V} = \left\{\mathbf{v}_j\right\}_{j=1}^M$, pretrained model $f_\theta$
	
\textbf{Output}: Model parameters $\theta$, hash center weights $\mathbf{w}$
	
\begin{algorithmic} %[1] enables line numbers
	
	    \STATE{Initialize $w_{i1}=w_{i2}=...=w_{ic_i}=\frac{1}{c_i}$ }
	    \FOR{$i$=$1$ to $N$}
	    \STATE{Forward Model $f_\theta$ to learn the hash codes $\{\mathbf{b}_i\}$} \\
	    %\STATE{Fixed $f_\theta$}
	    \STATE{Compute Hamming distance $d_{i}$ by Eq. (\ref{eq2})}
	    \STATE{Calculate weights $\mathbf{w}_i$ by projected gradient descent}
	    \ENDFOR
	    
	    \STATE{Compute loss $\mathcal{J}$ by Eq. (\ref{eq9})}
	    \STATE{Update $f_\theta$ by back-propagation with fixed $\mathbf{w}$}
	   
\end{algorithmic}
\label{alg:algorithm}
\end{algorithm}

\subsection{Alternative Optimization}
In order to automatically solve center weight $\mathbf{w}$ and perform model training, we propose a two-step alternative optimization approach. In the first step, parameters of model $f_\theta$ are frozen and we compute the hash codes $\{\mathbf{v}_{i}\}_{i=1}^{N}$ via $f_{\theta}$ and Hamming distance $\{\mathbf{d}_{i}\}_{i=1}^{N}$, we then utilize project gradient descent~\cite{pgd2013} to optimize $\{\mathbf{w}_{i}\}_{i=1}^{N}$. In the second step, update $f_\theta$ by backpropagation with fixed $\mathbf{w}$. We briefly summarize the major workflow of our proposed ICS method in Algorithm~\ref{alg:algorithm}. The solving process of $\mathbf{w}$ is as follows.

The gradient vector of the objective function $\mathcal{J}$ with respect to $\mathbf{w}_{i}$ can be derived as:
\begin{equation}
\label{pgd1}
\nabla w_{ij}=-w_{ij}^{t-1} \frac{1}{1+\exp (w_{ij}^{t-1} d_{ij})}+\lambda(1+\log(w_{ij}^{t-1})),
\end{equation}
where $w_{ij}^{t}$ denotes the value of parameter $w_{ij}$ at iteration t. Hence, we can find the optimum value of $w_{ij}$ by gradient descent:
\begin{equation}
\label{pgd2}
\begin{split}
\overline{w}_{ij}^t =& w_{ij}^{t-1}-\eta \nabla w_{ij} \\
=& w_{ij}^{t-1}-\eta [-w_{ij}^{t-1} \frac{1}{1+\exp (w_{ij}^{t-1} d_{ij})}+\lambda(1+\log(w_{ij}^{t-1}))],
\end{split}
\end{equation}
where $\eta$ denotes the learning rate. In all experiments, $\eta$ is set to 0.1. After obtaining solution $\overline{\mathbf{w}}_i^t$, in order to compute the Euclidean projection of $\overline{\mathbf{w}}_i^t$ onto the probability simplex, the optimization problem is defined as follows:
\begin{equation}
\centering
\label{pgd3}
\begin{split}
    \mathbf{w}_i^t=\min_{\mathbf{w}_i^t}\frac{1}{2}|\mathbf{w}_i^t-\overline{\mathbf{w}}_i^t|, \\
    s.t. \sum_{j=1}^{c_i} w_{ij}^t=1, \\
    w_{ij} \geq 0.
\end{split}
\end{equation}
The solution~\cite{pgd2013} of Eq.~\ref{pgd3} is presented in Algorithm~\ref{algorithm-pgd}.
\begin{algorithm}[tb]
	\caption{Euclidean projection of $\overline{\mathbf{w}}_i^t$ onto the probability simplex}
		
	\textbf{Input:} $\overline{\mathbf{w}}_i^t \in \mathbb{R}^{c_i}$
	
	\textbf{Output}: $\mathbf{w}_i^t s.t. \mathbf{w}_{ij}^t$ = max$\left\{\overline{\mathbf{w}}_{ij}^t + \xi, 0 \right\}, j = 1,2, ..., c_i$
	
	\begin{algorithmic}[1]

	    \STATE{Sort $\overline{\mathbf{w}}_i^t$ into $\mathbf{q}: q_1 \geq q_2 \geq ... \geq q_{c_i} $}
	  
	    \STATE{Find $\rho$ = max$\left\{1 \leq j \leq c_i: q_j + \frac{1}{j}(1- \sum_{i=1}^{j}q_i) > 0 \right\}$} 

	    \STATE{Define $\xi = \frac{1}{\rho}(1-\sum_{i=1}^{\rho}q_i)$}

	\end{algorithmic}
\label{algorithm-pgd}
\end{algorithm}

\section{Experiments}
\label{sec:experiment}
We first perform extensive experiments to evaluate our ICS against several state-of-the-art hashing methods on three multi-label datasets, NUS-WIDE~\cite{nus-wide2009} and MS COCO~\cite{coco2014}. %FLICKR25K~\cite{flickr2008}. 
Afterwards, we visualize the convergence behavior of solving the optimization via project gradient descent. Finally, we conduct ablation studies to analyze the effectiveness of the learned hash center weights.

\subsection{Datasets}
The following three benchmark datasets are used in our experiments and their statistics are summarized in Table ~\ref{table:dataset}.

\textbf{NUS-WIDE} contains about 0.27M images in 81 categories~\cite{nus-wide2009}. Each image belongs to more than one label. We follow similar experimental protocols as CSQ~\cite{CSQ2020}, we randomly sample 10,000 images as training set, 5,000 images as queries, the remaining images are used as database.

\textbf{MS COCO} is an image recognition, segmentation, and captioning dataset, which contains 80 common object categories with 82,783 training images and 40,504 validation images~\cite{coco2014}. We randomly sample 5,000 images as queries, with the rest images are used as the database; furthermore, we randomly sample 10,000 images from the database as training set.

\begin{table}
\caption{Statistics of the datasets used for training and evaluation.}

\begin{center}
\begin{tabular}{c|cccc}

\hline
Dataset          & \#Train    & \#Test     & \#Retrieval    & \#Label  \\
         
\hline 
NUS-WIDE         & 10,000     & 2,040      & 149,685        & 2.7       \\
        
MS COCO          & 10,000     & 5,000      & 112,217        & 2.9       \\
\hline

\end{tabular}
\end{center}

\label{table:dataset}
\end{table}

\subsection{Baselines and Evaluation Metrics}
We compare the retrieval performance of our proposed ICS with the state-of-the-art hashing methods, %The first type is the methods represented by 
CSQ~\cite{CSQ2020}, HashNet~\cite{hashnet2017}, 
% which do not consider multi-labels in image retrieval. Other baselines belonging to this type include 
LSH~\cite{LSH1999}, SH~\cite{SH2009}, BRE~\cite{BRE2009}, KSH~\cite{KSH2012}, ITQ~\cite{ITQ2013}, SDH~\cite{SDH2015}, CNNH~\cite{CNNH2014}, DNNH~\cite{DNNH2015}, DHN~\cite{DHN2016}, DCH~\cite{DCH2018}. We compute the mean Average Precision@5,000 (mAP@5,000) with top 5,000 searched results on NUS-WIDE and MS COCO datasets.

\begin{table*}[t]
\caption{Comparison in mAP of Hamming Ranking using AlexNet backbone for different bits on NUS-WIDE and MS COCO datasets. \\ * represents our reproducible results based on author's code.}
\begin{center}

\begin{tabular}{p{2.2cm}|p{1.6cm}p{1.6cm}p{1.6cm}|p{1.6cm}p{1.6cm}p{1.6cm}}
\hline
       
\multirow{2}{30pt}{\centering Method}   & \multicolumn{3}{c|}{NUS-WIDE (mAP@5,000)} & \multicolumn{3}{c}{MS COCO (mAP@5,000)} \\
    
\cline{2-7}
& 16 bits     & 32 bits    & 64 bits    & 16 bits    & 32 bits     & 64 bits \\
        
\hline
LSH~\cite{LSH1999}            & 0.328   & 0.422  & 0.500  & 0.459  & 0.485   & 0.584 \\
SH~\cite{SH2009}              & 0.405   & 0.420  & 0.410  & 0.495  & 0.507   & 0.510 \\
BRE~\cite{BRE2009}            & 0.502   & 0.529  & 0.554  & 0.592  & 0.622   & 0.633 \\
KSH~\cite{KSH2012}            & 0.356   & 0.332  & 0.336  & 0.521  & 0.534   & 0.536 \\
ITQ~\cite{ITQ2013}            & 0.508   & 0.542  & 0.561  & 0.581  & 0.624   & 0.657 \\
SDH~\cite{SDH2015}            & 0.475   & 0.554  & 0.581  & 0.554  & 0.564   & 0.579 \\

\hline
CNNH~\cite{CNNH2014}          & 0.569  & 0.582  & 0.599  & 0.564  & 0.574  & 0.567 \\
DNNH~\cite{DNNH2015}          & 0.597  & 0.615  & 0.638  & 0.593  & 0.603  & 0.609 \\
DHN~\cite{DHN2016}            & 0.637  & 0.663  & 0.669  & 0.677  & 0.701  & 0.694 \\
HashNet~\cite{hashnet2017}    & 0.662  & 0.698  & 0.716  & 0.687  & 0.718  & 0.736 \\
% DCH~\cite{DCH2018}          & 0.740  & 0.772  & 0.712  & 0.701  & 0.758  & 0.701 \\
CSQ*~\cite{CSQ2020}            & 0.744  & 0.785  & 0.789  & 0.635  & 0.708  & 0.748 \\

\hline 
\textbf{Ours}   & \textbf{0.755}  & \textbf{0.789}   & \textbf{0.794} & \textbf{0.699}  & \textbf{0.736}  & \textbf{0.764}   \\
\hline
    
\end{tabular}
\end{center}
\label{table:alexnet}
\end{table*}

\begin{table*}[tb]
\caption{Comparison in mAP of Hamming Ranking using ResNet-50 backbone for different bits on NUS-WIDE and MS COCO datasets. \\ * represents our reproducible results based on author's code.}
\begin{center}

\begin{tabular}{p{2.2cm}|p{1.6cm}p{1.6cm}p{1.6cm}|p{1.6cm}p{1.6cm}p{1.6cm}}
\hline
       
\multirow{2}{30pt}{\centering Method}     & \multicolumn{3}{c|}{NUS-WIDE (mAP@5,000)} & \multicolumn{3}{c}{MS COCO (mAP@5,000)} \\
    
\cline{2-7}
& 16 bits     & 32 bits    & 64 bits    & 16 bits    & 32 bits     & 64 bits \\
        
\hline
BRE~\cite{BRE2009}            & 0.485   & 0.525  & 0.544  & 0.592  & 0.622   & 0.634 \\
KSH~\cite{KSH2012}            & 0.394   & 0.407  & 0.399  & 0.521  & 0.534   & 0.536 \\
ITQ~\cite{ITQ2013}            & 0.435   & 0.435  & 0.435  & 0.566  & 0.562   & 0.502 \\
SDH~\cite{SDH2015}            & 0.575   & 0.590  & 0.613  & 0.554  & 0.564   & 0.580 \\

\hline
CNNH~\cite{CNNH2014}          & 0.655  & 0.659  & 0.647  & 0.599  & 0.617  & 0.620 \\
DNNH~\cite{DNNH2015}          & 0.703  & 0.738  & 0.754  & 0.644  & 0.651  & 0.647 \\
DHN~\cite{DHN2016}            & 0.719  & 0.731  & 0.745  & 0.719  & 0.731  & 0.745 \\
HashNet~\cite{hashnet2017}    & 0.757  & 0.775  & 0.790  & 0.745  & 0.773  & 0.788 \\
DCH~\cite{DCH2018}            & 0.773  & 0.795  & 0.818  & 0.759  & 0.801  & 0.825 \\
CSQ*~\cite{CSQ2020}           & 0.789  & 0.829  & 0.831  & 0.772  & 0.849  & 0.866 \\

\hline 
\textbf{Ours}   & \textbf{0.809}  & \textbf{0.833}   & \textbf{0.833} & \textbf{0.838}  & \textbf{0.865}  & \textbf{0.880}   \\
\hline

\end{tabular}
\end{center}
\label{table:resnet-50}
\end{table*}

\begin{figure}
\begin{minipage}{\linewidth}
\centering
\includegraphics[scale=0.471]{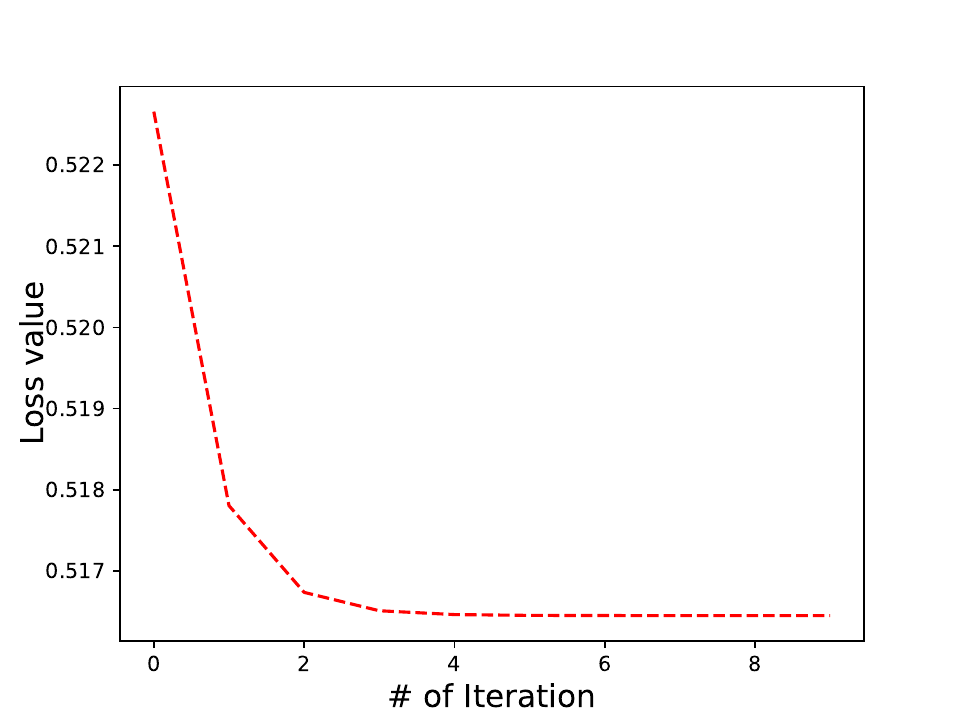}
\end{minipage}
		
\caption{The loss curve when using of project gradient descent to solve the weight values of hash centers. The learning rate is set to 0.1.}
\label{fig:loss}
\end{figure}

% ----------------------------------------------------------------- %
\subsection{Implementation Details}
We implement the ICS method on the Pytorch framework~\cite{Pytorch2017}, and adopt AlexNet~\cite{AlexNet2012} and ResNet-50~\cite{ResNet2016} 
pre-trained on ImageNet as backbone architectures to evaluate the performance of our proposed method. We use the standard data argumentation strategies such as random horizontal flip and random crop. Each image is resized to 256$\times$256 and then is cropped to 224$\times$224. We apply Adam optimizer with $\beta_1=0.9$ and $\beta_2=0.99$ to optimize model parameters, and set weight decay to 0.9 and batch size to 64. The neural network is trained for 90 epochs where the learning rate starts from 0.0001 and is divided by 10 every 30 epoch.

% ----------------------------------------------------------------- %
\subsection{Experimental Results}
As illustrated in Table~\ref{table:alexnet}, our method using AlexNet backbone increases the performance by 1.1\%, 0.4\%, 0.5\% for 16 bits, 32 bits and 64 bits on the NUS-WIDE, and yields an improvement of 6.4\%, 2.8\%, 1.6\%  for 16 bits, 32 bits, and 64 bits on the MS COCO. The results in Table~\ref{table:resnet-50} achieve absolute improvements of 2\% and 6.6\% for 16 bits on the two datasets, respectively. The above impressive results clearly demonstrate that our method achieves state-of-the-art performance by considering the relationship between the hash center weights and the proportion of foreground instances in the image.

\textbf{Convergence of Project Gradient Descent}
We would like to study the convergence behaviour of the adopted projected gradient descent for automatically learning the weight values of hash centers. Figure~\ref{fig:loss} shows that the loss converges within 10 iterations or even fewer, which demonstrates that PGD can solve the optimization problem efficiently. The extra computational cost of PGD is relatively small, the time consuming is about 8 ms.

\begin{figure*}[t]
	\centering
		%\begin{minipage}{\linewidth}\centering
		%	\includegraphics[scale=0.7]{img/visualization.pdf}
		%\end{minipage}
		
		\begin{minipage}{\linewidth}\centering
			\includegraphics[scale=0.72]{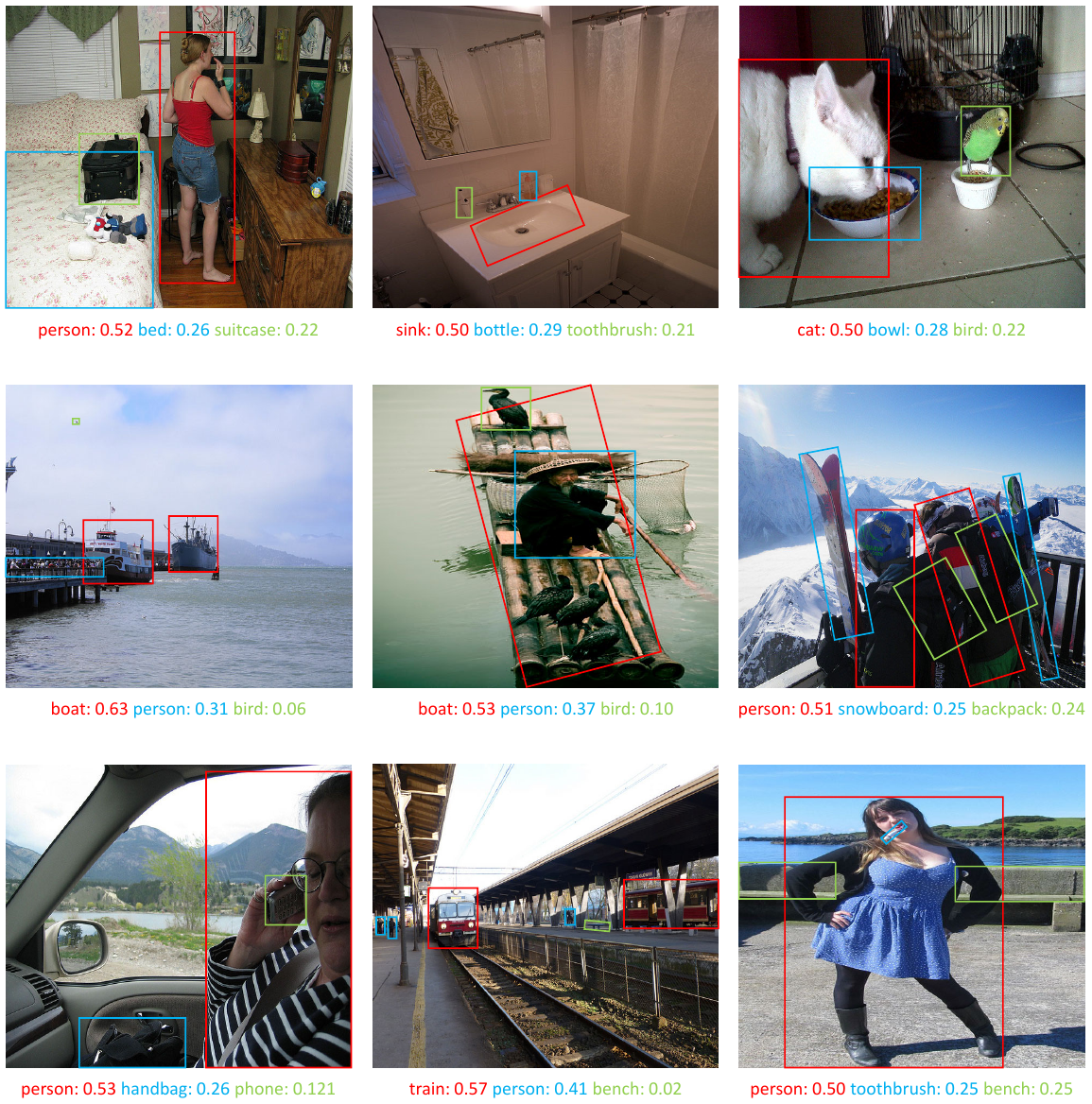}
		\end{minipage}
		
\caption{Visualization of hash center weights on MS COCO dataset trained by AlexNet in 64 bits.}

\end{figure*}\label{fig:visualization}

% ----------------------------------------------------------------- %
\subsection{Ablation Study}
We analyse the effectiveness of our proposed ICS in the following three aspects.

\begin{table*}
\caption{Comparision in mAP of E-WCS and ICS using AlexNet backbone for different bits on NUS-WIDE and MS COCO datasets.}
\begin{center}

\begin{tabular}{p{2.2cm}|p{1.5cm}p{1.5cm}p{1.5cm}|p{1.5cm}p{1.5cm}p{1.5cm}}
\toprule

\multirow{2}{30pt}{\centering Method} &   \multicolumn{3}{c|}{NUS-WIDE (mAP@5,000)} & \multicolumn{3}{c}{MS COCO (mAP@5,000)} \\
\cline{2-7}
& 16 bits    & 32 bits          & 64 bits      & 16 bits    & 32 bits    & 64 bits    \\

\hline
CSQ~\cite{CSQ2020}   & 0.744  & 0.785  & 0.789  & 0.635  & 0.708  & 0.748 \\

\hline
E-WCS              & 0.743    & 0.777          & 0.779       & 0.642   & 0.717     & 0.741    \\

\textbf{ICS}  & \textbf{0.755}  & \textbf{0.789}  & \textbf{0.794}  & \textbf{0.699}  & \textbf{0.736}  & \textbf{0.764}  \\
    
\bottomrule
  
\end{tabular}
\end{center}
\label{tab:equal}
\end{table*}

\subsubsection{Equal Center Weights} 
We compare our proposed ICS with its variant, Equal Weighted Central Similarity (E-WCS). Different from CSQ~\cite{CSQ2020} using single centroid as learning target, E-WCS uses multiple hash centers as learning target and fixes their weights with equal values, $\frac{1}{c_i}$. As illutrated in Table~\ref{tab:equal}, the average performance of E-WCS is inferior to CSQ, and ICS achieves higher mAP than E-WCS and CSQ for different bits on NUS-WIDE and MS COCO datasets. In addition, The results demonstrate the necessity of automatically learning the hash center weights, rather than fixing them with equal values or learning with single centroid.

\begin{figure*}[t]
\centering
\small
\begin{tabular}{c@{ }@{ }c@{ }@{ }c@{ }@{ }c}

\begin{minipage}{0.32\linewidth}\centering
    \includegraphics[scale=0.4]{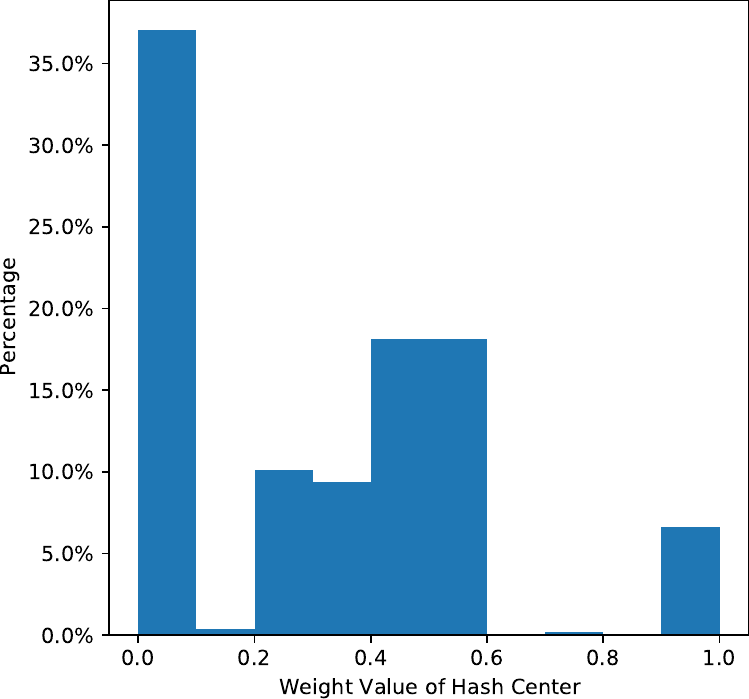}\\
    (a) $\lambda$=10, mAP=0.742, var=0.0650
\end{minipage} &

\begin{minipage}{0.32\linewidth}\centering
    \includegraphics[scale=0.4]{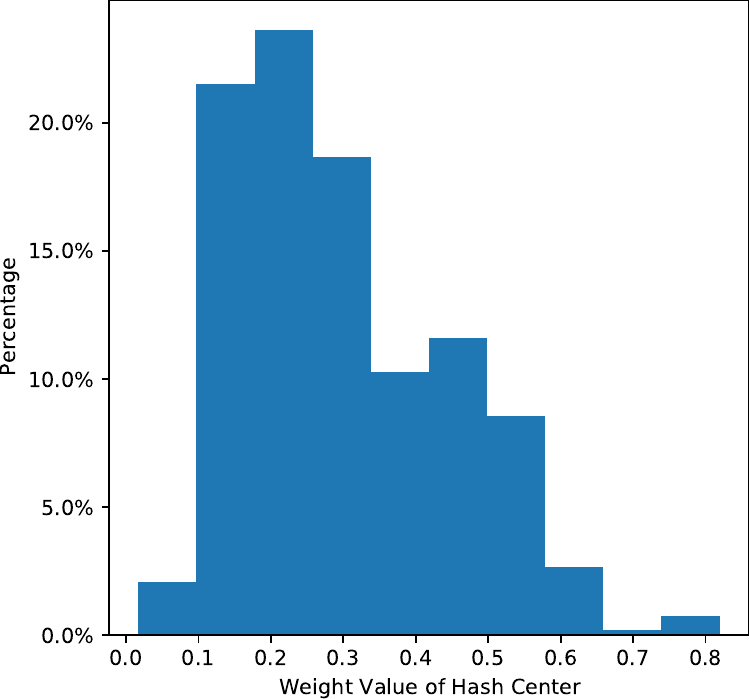}\\
    (b) $\lambda$=0.1, mAP=0.758, var=0.0214
\end{minipage} &

\begin{minipage}{0.32\linewidth}\centering
    \includegraphics[scale=0.4]{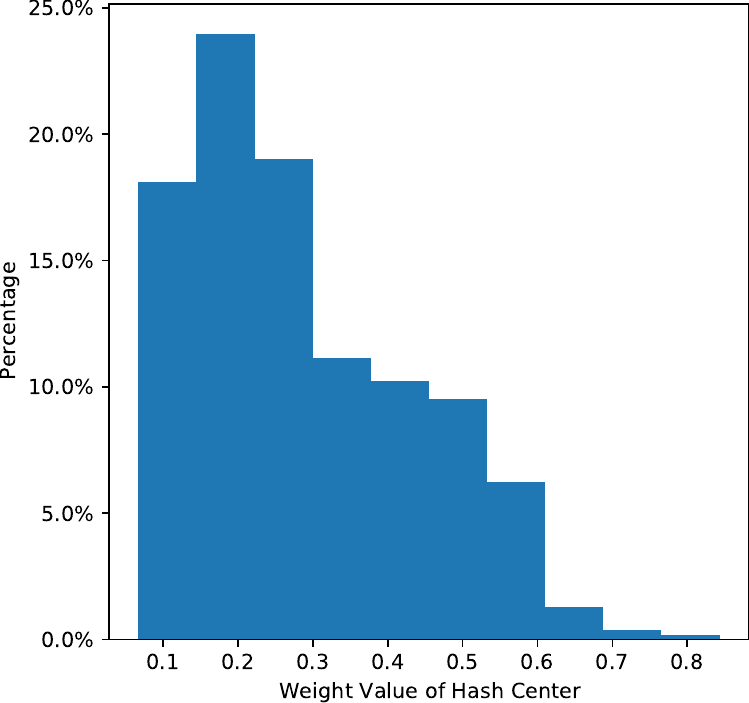}\\
    (c) \textbf{$\lambda$=0.01, mAP=0.764, var=0.0211}
\end{minipage} &

\end{tabular}
\caption{Visualization of weight distribution of hash centers on MS COCO training set with AlexNet in 64 bits varying trade-off parameter $\lambda$. The ``var'' denotes the variance of weights.} 
\label{fig:weight_distribution_coco_64bits}
\end{figure*}

\begin{figure}[t]

		\begin{minipage}{0.92\linewidth}
		\centering
			\includegraphics[scale=0.49]{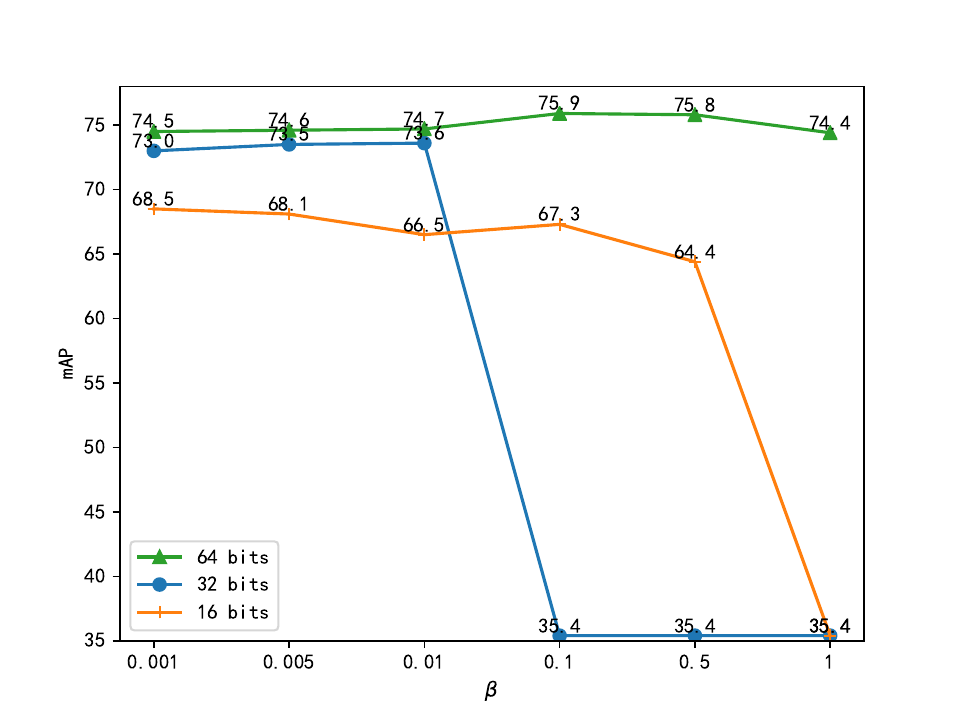}
		\end{minipage}
		
		\caption{Comparision in mAP using AlexNet backbone for different bits on MS COCO by varying coefficient $\beta$.}
		\label{fig:parameter_study_beta}
\end{figure}

\begin{figure}[t]

	\begin{minipage}{0.8\linewidth}
	\includegraphics[scale=0.49]{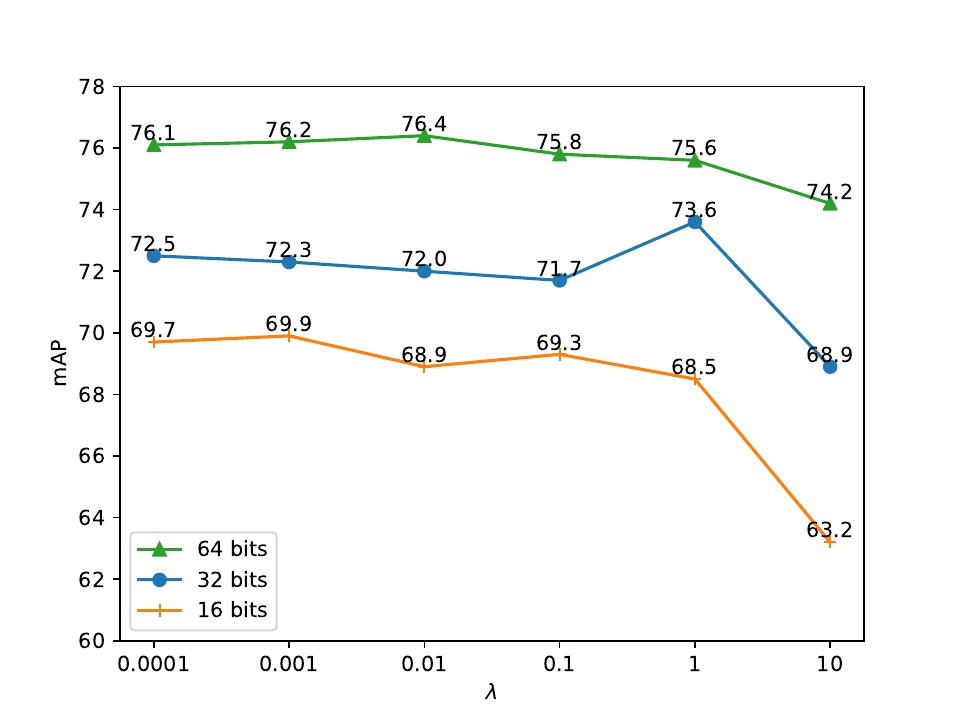}
	\end{minipage}
		
\caption{Comparision in mAP using AlexNet backbone for different bits on MS COCO by varying coefficient $\lambda$.}
\label{fig:parameter_study_lambda}
\end{figure}

\subsubsection{Parameter Analysis}
We firstly tune the parameter $\beta$ with fixed $\lambda=1$, and then choose the optimal parameter $\beta$ for different bits to adjust $\lambda$.
We construct experiments on the analysis of hyper-parameters using AlexNet backbone on Ms COCO dataset, and the chosen hyper-parameters are directly applied to all the other datasets.

\textbf{Parameter $\bm{\beta}$}. The $\beta$ is a smooth parameter on the Sigmoid function Eq.~\ref{eq4} to control its bandwidth. As shown in Figure \ref{fig:parameter_study_beta}, When $\beta$ is 0.001, 0.01, 0.1, the model achieves the best performance on 16, 32, and 64 bits respectively. We can conclude that smaller bits achieve the best results with smaller $\beta$, % and larger $\beta$ will crash the model. 
For image retrieval in 64 bits, the distance between each hash code and its corresponding hash centers is larger than lower bits. Therefore, larger $\beta$ makes the curve of Sigmoid function more smoother to prevent the gradient from disappearing, which is consistent with the result of HashNet~\cite{hashnet2017}.

\textbf{Parameter $\bm{\lambda}$}. The $\lambda$ is used to control the maximum entropy regularization. We choose the value of $\lambda$ from $\{10, 1, 0.1, 0.01, 0.001, 0.0001\}$ and compare their performance on MS COCO dataset in 16, 32 and 64 bits. 
% From Figure \ref{fig:parameter_study_lambda}, we find that our proposed ICS is not very sensitive to the value of $\lambda$ when $\lambda \leq 1$. 
As discussed in Lemma~\ref{lemma} and illutrated in Figure~\ref{fig:parameter_study_lambda}, larger $\lambda$ makes one hash center dominate the loss resulting in poor retrieval performance. The results also demonstrate that the proposed maximum entropy regularization is not very sensitive to the value of $\lambda$ when $\lambda \leq 1$.

\subsubsection{Visualization of Center Weights}
We evaluate the model on the MS COCO dataset for visualizing the values of weights. Figure~\ref{fig:visualization} shows the weight value of hash center corresponding to each foreground instance belonging to different categories. For example, in the last images in the lower right corner, the center weights of ``person'', ``toothbrush'' and ``bench'' are 0.50, 0.25 and 0.25, respectively. The visualization results illustrate that the weight values learned by ICS can well reflect the proportion of each foreground instance in the image. Moreover, although the ``toothbrush'' occupy a smaller region than ``bench'' in the image, they have same values of center weights, which demonstrate that our proposed ICS can guides the model focusing on more semantic and salient regions to improve the retrieval performance. 

In addition, we also visualize the distribution of center weights. As shown in Figure~\ref{fig:weight_distribution_coco_64bits}, our method achieves the best performance $0.764$ where the variance of weights is smaller. Compared with CSQ~\cite{CSQ2020} in which all the weight values are equal to 1, our results demonstrate that the more uniform distribution of center weights is conducive to model learning for multi-label image retrieval.

\section{Conclusion}
In this paper, we propose to automatically learn the hash center weights by an alternative optimization approach for multi-label image retrieval. The maximum entropy regularizer prevents one hash center from dominating the loss function, and guides the model to identify the relative importance of hash centers. Our method is also an instance-aware method that the learned hash center weights can well reflect the proportion of foreground objects in the image. Experiments on two datasets validate the effectiveness of our proposed method. Beyond the multi-label image retrieval task, our method has great potential on other tasks, which needs to be further explored.

% \clearpage
% \newpage

{\small
\bibliographystyle{ieee_fullname}
\bibliography{egbib}
}

\end{document}